\documentclass{article}
\usepackage{graphicx}
\usepackage{textcomp}
\usepackage{stfloats}
\usepackage{url}
\usepackage{verbatim}
\usepackage{graphicx}
\usepackage{cite}
\usepackage{booktabs,multirow}
\usepackage{makecell}
\usepackage{bigstrut}
\usepackage{array}
\usepackage{amsmath}
\usepackage{amsmath,amsfonts}
\usepackage{algorithmic}
\usepackage{algorithm}
\usepackage{array}
\usepackage{float}
\usepackage{graphicx}
\usepackage{url}

\usepackage[preprint]{neurips_2023}

\usepackage[utf8]{inputenc} 
\usepackage[T1]{fontenc}    
\usepackage{hyperref}       
\usepackage{url}            
\usepackage{booktabs}       
\usepackage{amsfonts}       
\usepackage{nicefrac}       
\usepackage{microtype}      
\usepackage{xcolor}         

\title{SAMN: A Sample Attention Memory Network Combining SVM and NN in One Architecture}

\author{
  Qiaoling Yang \\
  Department of Automation, \\
  Xiamen University,\\
  Xiamen, China \\
  \texttt{qiaolingyang@foxmail.com} 
  \And
  Linkai Luo\thanks{Corresponding Author. This work was supported in part by the China Natural Science Foundation under Grant 62171391.}\\
  Department of Automation, \\
  Xiamen University,\\
  Xiamen, China \\
  \texttt{luolk@xmu.edu.cn} \\
  \And
  Haoyu Zhang \\
  Department of Automation, \\
  Xiamen University,\\
  Xiamen, China \\
  \texttt{2278992156@qq.com} \\
  \And
  Hong Peng \\
  Department of Automation, \\
  Xiamen University,\\
  Xiamen, China \\
  \texttt{penghong@xmu.edu.cn} \\
  \And
  Ziyang Chen \\
  Department of Automation, \\
  Xiamen University,\\
  Xiamen, China \\
  \texttt{ziyang.chan@foxmail.com} \\
}

\begin{document}

\maketitle

\begin{abstract}
Support vector machine (SVM) and neural networks (NN) have strong complementarity. SVM focuses on the inner operation among samples while NN focuses on the operation among the features within samples. Thus, it is promising and attractive to combine SVM and NN, as it may provide a more powerful function than SVM or NN alone. However, current work on combining them lacks true integration. To address this, we propose a sample attention memory network (SAMN) that effectively combines SVM and NN by incorporating sample attention module, class prototypes, and memory block to NN. SVM can be viewed as a sample attention machine. It allows us to add a sample attention module to NN to implement the main function of SVM. Class prototypes are representatives of all classes, which can be viewed as alternatives to support vectors. The memory block is used for the storage and update of class prototypes. Class prototypes and memory block effectively reduce the computational cost of sample attention and make SAMN suitable for multi-classification tasks. Extensive experiments show that SAMN achieves better classification performance than single SVM or single NN with similar parameter sizes, as well as the previous best model for combining SVM and NN. The sample attention mechanism is a flexible module that can be easily deepened and incorporated into neural networks that require it.
\end{abstract}

\section{Introduction}
Support vector machine (SVM) \cite{1995support} and neural networks (NN) \cite{rosenblatt1958perceptron,graves2012long,he2016deep,huang2017densely,krizhevsky2017imagenet} are two of the most famous machine learning models that have profoundly influenced current machine learning. SVM and NN have strong complementarity in their principles. SVM focuses on the inner product operation among samples and does not consider the operation among internal features of a sample. On the other hand, NN focuses on the operation among internal features of a sample and does not consider the operation among samples. In recent years, graph neural networks \cite{kipfsemi,velivckovicgraph,chen2020simple,zhao2022multi} have considered the operation among samples on graph data where the relationship among samples is given by a graph. However, NN still does not consider the operation among samples for non-graph data. Due to these characteristics, each of them has its limitations. For example, SVM suffers from high time costs for large-scale classification tasks because it requires solving a large number of quadratic programming sub-problems in the training process. Furthermore, SVM needs to save support vectors for the test, which causes inconvenience when there are many support vectors. These limitations limit the application of SVM. In comparison, NN has greatly improved its time costs with the famous backpropagation (BP) algorithm \cite{rumelhart1986learning}. However, NN suffers from the vanishing gradient problem and can fall into local minima or saddle points \cite{choromanska2015loss,dauphin2014identifying}. Additionally, NN does not perform well when the operations among samples have an important impact on the results, as it does not consider the operations. These limitations still exist in current deep learning methods based on NN.

Combining SVM and NN within one architecture is a promising and attractive direction due to their strong complementarity. SVM is a constrained optimization problem while NN is an unconstrained optimization problem. Thus, it is not easy to combine them in one architecture, and there are many challenges. Existing work in this area is limited and can be divided into two categories. The first category involves using NN to extract advanced features and then feeding them to SVM for classification. Huang et al. \cite{huang2006large} used a convolutional network to extract advanced features, and then fed the extracted features to SVM. Ghanty et al. \cite{ghanty2009neurosvm} replaced a single network with multiple neural networks to extract more valuable features. In this category, NN and SVM are trained independently in stages, making it impossible to modify the trained NN by the following SVM. Therefore, the training of NN and SVM is not truly unified in one architecture. The second category also uses NN to extract advanced features, but SVM is transformed into an unconstrained optimization problem so that NN and SVM can be trained within one architecture by the BP algorithm. Li et al.'s deep neural mapping support vector machines (DNMSVM) \cite{li2017deep} belong to this category. DNMSVM achieves better performance than \cite{huang2006large} and \cite{ghanty2009neurosvm}. However, updating the parameters of NN and SVM alternately in DNMSVM leads to the loss of the maximum margin of SVM. Additionally, DNMSVM is not suitable for multi-classification tasks.

To summarize, the current work on combining SVM and NN lacks true integration. To address these issues, we propose a sample attention memory network (SAMN), which incorporates sample attention module, class prototypes, and a memory block to effectively combine SVM and NN. Firstly, we need to conduct a deeper investigation of SVM so that SVM and NN can be effectively combined. In fact, SVM can be viewed as a sample attention machine. In SVM with a linear kernel, the classification function of a test sample $\boldsymbol{x}$ can be simplified as
\begin{equation}
y(\boldsymbol{x}) = sign( {\sum\limits_{i = 1}^{l}{y_{i}\alpha_{i}^{*}{<\boldsymbol{x}_{i}}, \boldsymbol{x}}}> ),
\end{equation}
where $\boldsymbol{x}_{i},\boldsymbol{x} \in \boldsymbol{R}^{n}$ indicate the input feature vector of samples, $y_{i} \in \left\{ 1, - 1 \right\}$ are the class label of training samples $\boldsymbol{x}_i$, $l$ is the number of training samples, $\alpha_{i}^{*}$ is an optimal solution of SVM model, $<\cdot,\cdot>$ denotes the inner operation, $sign(\cdot)$ is the symbolic function. In fact, $y(\boldsymbol{x})$ can be viewed as a sample attention function where $\alpha_{i}^{*}$ is the attention coefficient of training sample $\boldsymbol{x}_i$. In addition, $\alpha_{i}^{*}$ is from the inner product operations among training samples since $\alpha_{i}^{*}$ is an optimal solution of the SVM model. Therefore, SVM can be seen as a sample attention machine. It allows us to add sample attention module to NN to achieve the combination of SVM and NN. Secondly, the computational cost in the sample attention module is high when the number of training samples is large. Inspired by the memory mechanism of recurrent neural networks (RNN) \cite{graves2013speech,sutskever2014sequence,liang2015recurrent}, we introduce class prototypes and memory block to reduce the computational cost. The class prototypes are used to replace support vectors, which record the learned information for each class. The memory block is used for the storage and update of class prototypes. 
 
The main contributions of this paper are as follows: (1) Propose a viewpoint of viewing SVM as a sample attention machine. (2) Introduce class prototypes and memory block to reduce the computational complexity of sample attention and make it applicable to multi-classification tasks. (3) Propose a SAMN that consists of sample attention module, class prototypes, memory block, as well as regular NN units, so that the complementary of SVM and NN is achieved. The experimental results show that SAMN outperforms a single SVM, a single NN with similar parameter size, and the previous best model in combining SVM and NN, achieving a new state-of-the-art result.

The remainder of the paper is organized as follows. We introduce related work in Section \ref{related work}. The details of SAMN will be introduced in Section \ref{samn}. The experimental results and analysis are in Section \ref{experiments}. Finally, the conclusions and the future work are presented in Section \ref{conclusion}.

\section{Related work}
\label{related work}
\paragraph{DNMSVM}
DNMSVM consists of a feature extraction module (FM) and a classification module (CM). The FM includes an input layer and multiple hidden layers, while the CM is essentially an SVM with only one binary unit in the output layer. DNMSVM feeds the features extracted by the last hidden layer of FM into the SVM, where the FM plays the role of a kernel function, with its kernel mapping being $\boldsymbol{\Phi}\left( {\boldsymbol{x}} \middle| \theta \right) = \boldsymbol{h}_{r}$, where $\boldsymbol{h}_j$ is the $j$-th hidden layer in the FM module, $\theta = \left\{ {\boldsymbol{W}}_{j},~\boldsymbol{b}_{j},~1 \leq j \leq r \right\}$. Let ${\boldsymbol{W}}_{r + 1}$ and ${\boldsymbol{b}}_{r + 1}$ be the parameters of the last hidden layer, representing the weight matrix and bias of the SVM in CM. DNMSVM reformulates the form of the SVM into an unconstrained optimization problem 
\begin{equation}
\label{loss}
{\min\limits_{{~\boldsymbol{W}}_{j},~\boldsymbol{b}_{j}, 1 \leq \mathit{j} \leq \mathit{r} + 1}{{\frac{1}{2}\left\| {\boldsymbol{W}}_{r + 1} \right\|}^{2} + C{\sum\limits_{i = 1}^{l}{{\lbrack\max( 1 - y_{i}\left( {\boldsymbol{W}}_{r + 1} \right.^{T}\Phi\left( \boldsymbol{x}_{i} \right) + b ),0)\rbrack}^{2},}}}}
\end{equation}
which is trained by gradient descent algorithm. DNMSVM can be viewed as a new type of general kernel learning method that can approximate any kernel SVM without using kernel tricks, as deep neural networks with sufficient capacity can approximate any kernel function \cite{hornik1991approximation}.

\paragraph{Attention mechanism}
The proposal of the attention mechanism originates from the simulation of human cognitive behavior, which focuses attention on important information and ignores irrelevant information during learning and cognition. The flexibility of the attention mechanism arises from its role as 'soft weight', which can automatically learn weight coefficients during network updates compared to fixed weights. The original attention mechanism involved only three inputs: Query ($\boldsymbol{Q}$), Key ($\boldsymbol{K}$), and Value ($\boldsymbol{V}$), and the outputs were generated by calculating attention among them. A common method for calculating attention is:
\begin{equation}
Attention\left( {\boldsymbol{Q},\boldsymbol{K},\boldsymbol{V}} \right) = Softmax\left( \frac{\boldsymbol{Q} \times \boldsymbol{K}^{T}}{\sqrt{d}}\right) \cdot \boldsymbol{V},
\end{equation}
where $d$ is the number of columns of the matrix $\boldsymbol{K}$, $Softmax(\cdot)$ is a normalization function. The calculation method of the similarity matrix is not limited to the dot product operation, and there are other calculation methods \cite{bahdanau2015neural,luong2015effective}.

\paragraph{Memory networks}
RNN was proposed by Werbos \cite{werbos1988generalization} in 1988. RNN is a type of neural network that processes sequential inputs. It can be seen as a network structure with a loop composed of connections among nodes, allowing the output of some nodes to continue to influence the input of subsequent nodes. It can be understood as a type of memory network used to store and transmit messages. The memory function of RNN is completed by the recurrent neurons, and the learning process can be represented by the recursive formula:
\begin{equation}
\boldsymbol{h}\left( \boldsymbol{x}_{t} \right) = f\left( {\boldsymbol{h}\left( \boldsymbol{x}_{t - 1} \right),\boldsymbol{x}}_{t} \right).
\end{equation}
Here, $\boldsymbol{x}_t$ denotes the input sequence at the $t$-th time step, and $\boldsymbol{h}(\cdot)$ represents the transfer of information. Memory information is learned through this recursive formula, where the output at the current moment is jointly determined by the input at all previous moments and the input at the current moment.
\section{Sample attention memory network}
\label{samn}
In this section, we will provide the details of SAMN, including its network structure, sample attention module, class prototypes, and memory block.
\subsection{Network structure}
\label{network structure}
\begin{figure}[H]
\centering
\includegraphics[width=5.5in,height=1.2in]{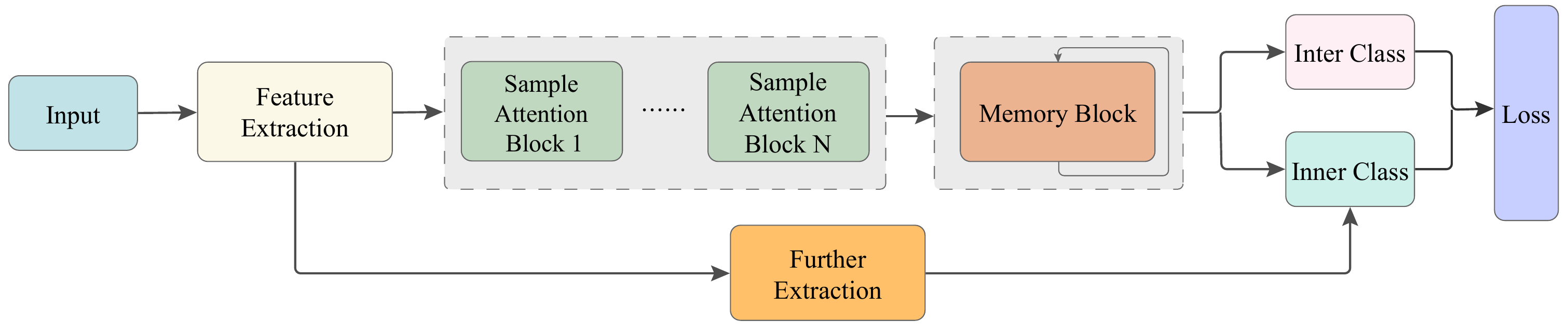}
\caption{The structure of SAMN.}
\label{fig1}
\end{figure}

The SAMN structure is shown in Figure \ref{fig1}. It is a fundamental framework. This framework includes feature extraction module, sample attention module, memory block, inter-class distance module, and inner-class distance module. First, input samples are extracted into higher-level feature representations through the corresponding feature extraction module. Then, the sample attention module uses the obtained high-level features to perform attention, which results in new feature representations that are computed through operations among samples. The sample attention can be stacked with multiple modules. To enable the network to be trained in batches, the memory block is used to remember the connections among samples and obtain the final class prototypes that are the representations of classes. The inter-class distance module and the inner-class distance module are used to construct the loss function of SAMN. In the inter-class distance module, the distance among different class prototypes is enlarged to increase the differences among final class prototypes. Before the inner-class distance module, the samples are further extracted into more refined high-level features. In the inner-class distance module, the similarity between the sample and the corresponding class prototype is calculated. The overall loss function of the network consists of two parts, including inter-class distance and inner-class distance, as shown in formula \eqref{formula_ex1}.
\begin{equation}
\label{formula_ex1}
L_{total} = L_{inter} + L_{inner}.
\end{equation}
\subsection{Sample attention module}
The sample attention mechanism is a good approach to consider the relationships among samples in NN. The attention mechanism enables the model to focus on certain information that needs to be focused on, such as the key features within the samples. In recent years, attention mechanisms have been widely used and have played an important role in many fields. However, existing attention mechanisms still only consider the information within the samples. To connect the relationships among samples, a sample attention mechanism is proposed where the mutual influence among samples is considered. The impact of different samples on the model's learning ability is different, and sample attention can help the model pay attention to more important samples.

\begin{figure}[H]
\centering
\includegraphics[width=4.8in,height=2.6in]{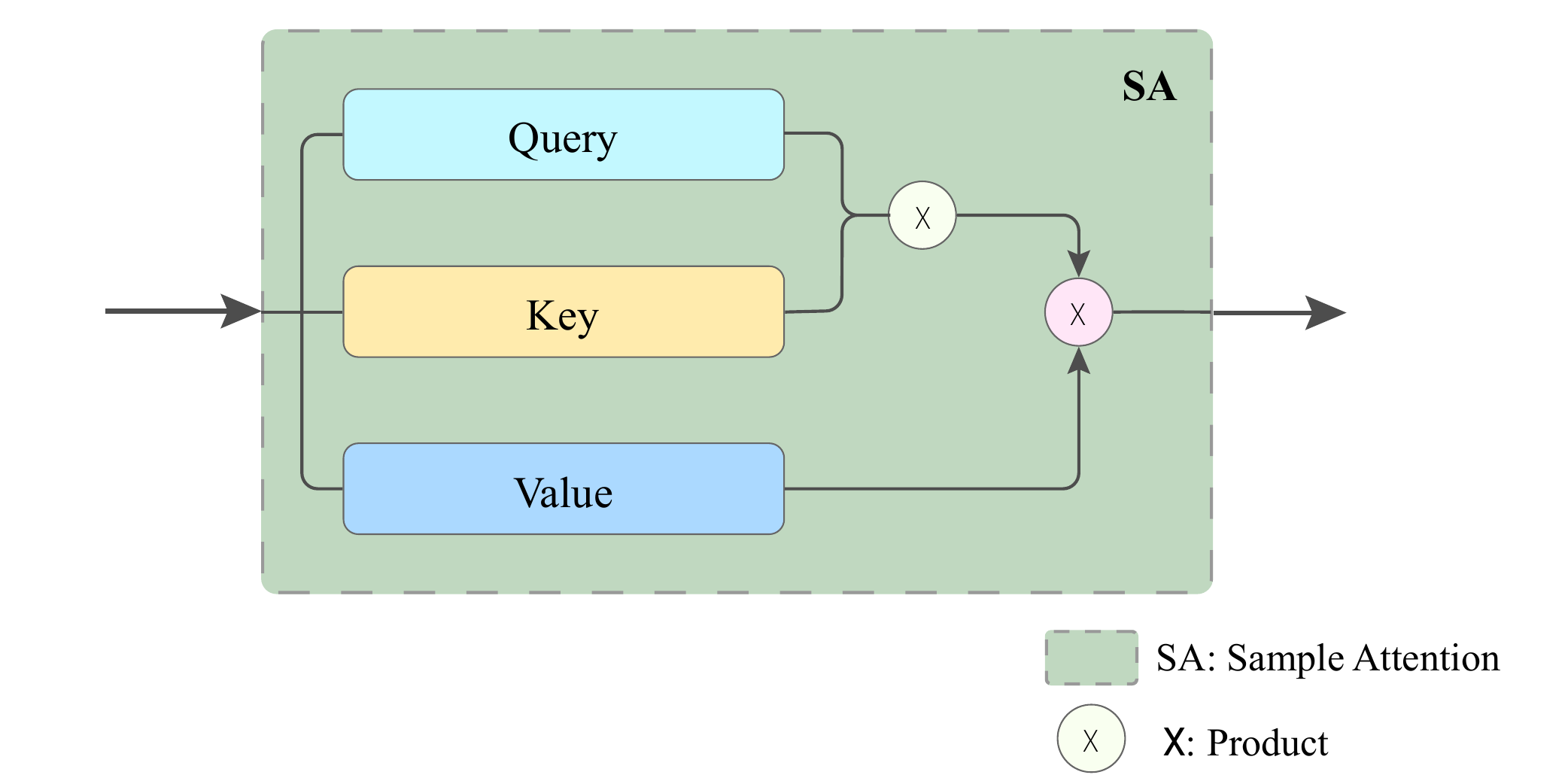}
\caption{A frame of sample attention mechanism. After performing the initial feature extraction, the high-level feature representation of the samples, Query ($\boldsymbol{Q}$), Key ($\boldsymbol{K}$), and Value ($\boldsymbol{V}$) are obtained.}
\label{fig2}
\end{figure}

The sample attention mechanism automatically learns the attention coefficients of samples by the operations among samples. A frame of sample attention mechanism is shown in Figure \ref{fig2}, which is similar to other attention mechanisms. The similarity matrix among samples is expressed as
\begin{equation}
\label{formula_ex2}
\boldsymbol{A} = \sigma\left( {\boldsymbol{Q} \times \boldsymbol{K}^{T}} \right).
\end{equation}

The shape of $\boldsymbol{Q}$, $\boldsymbol{K}$, and $\boldsymbol{V}$ are all $\left( {n \times d} \right)$, where $n$ is the number of samples and $d$ is the feature dimension of the samples. After calculating the similarity, the similarity matrix obtained has a shape of $\left( {n \times n} \right)$ and needs to be row-normalized. The Softmax function $\sigma( \cdot )$ is used to normalize the similarity matrix, making the sum of the similarities in each row equal to 1 and conforming to a probability distribution.

After obtaining the similarity matrix among samples, the sample features can be transformed by incorporating information from other samples, resulting in a new feature representation. The expression for the new sample feature is as follows:
\begin{equation}
\label{formula_ex3}
\boldsymbol{X} = \boldsymbol{A} \times \boldsymbol{V}.
\end{equation}
Sample attention is a powerful mechanism that enables the learning of relationships among samples on non-graph data. By using attention mechanisms, both explicit and intrinsic relationships among samples can be learned.
\subsection{Class Prototypes}
Machine learning models are commonly trained on large datasets in application tasks. However, the sample attention mechanism may incur high computational and memory costs since it needs to carry out the operations among all samples. Storing all samples is not practical, hence we propose using class prototypes to capture the crucial information learned from the samples. Class prototypes refer to representations of classes.

We simply use the mean of all the samples in that class as the class prototype. The class prototypes are expressed as 
\begin{equation}
\label{formula_ex3}
{{\overline{\boldsymbol{X}}^{c}}} = mean\left( \boldsymbol{X}^{c} \right),~c \in \left\{ {1,~2,~\ldots,C}\right\}.
\end{equation}
The variable $c$ represents the category to which a sample belongs, $C$ represents the total number of categories, and $\boldsymbol{X}^c$ represents the samples belonging to category $c$. The class prototypes $\boldsymbol{X}$ record all learned information of classes from all samples.

\subsection{Memory block}
To speed up training time and save computational memory, neural networks are usually trained in batches. It means that the learning of class prototypes cannot be based on all samples, but only on the current batch of samples. To solve this problem, we propose a memory block to help the network to be trained.

The memory block is used to store class prototypes so that previous sample information can be obtained during batch training. To achieve this goal, we consider using an RNN structure to handle temporal dependencies. The learning of class prototypes is based on all sample information, but the data must be divided into batches during training. We assume that these batches are interdependent because class prototypes must be generated from all samples. Therefore, in order to learn class prototypes in each batch, it is necessary to consider the information on class prototypes learned in the previous batches. We use a memory block to autonomously learn and adjust the information of final class prototypes.

\begin{figure}[H]
\centering
\includegraphics[width=4.8in,height=2.5in]{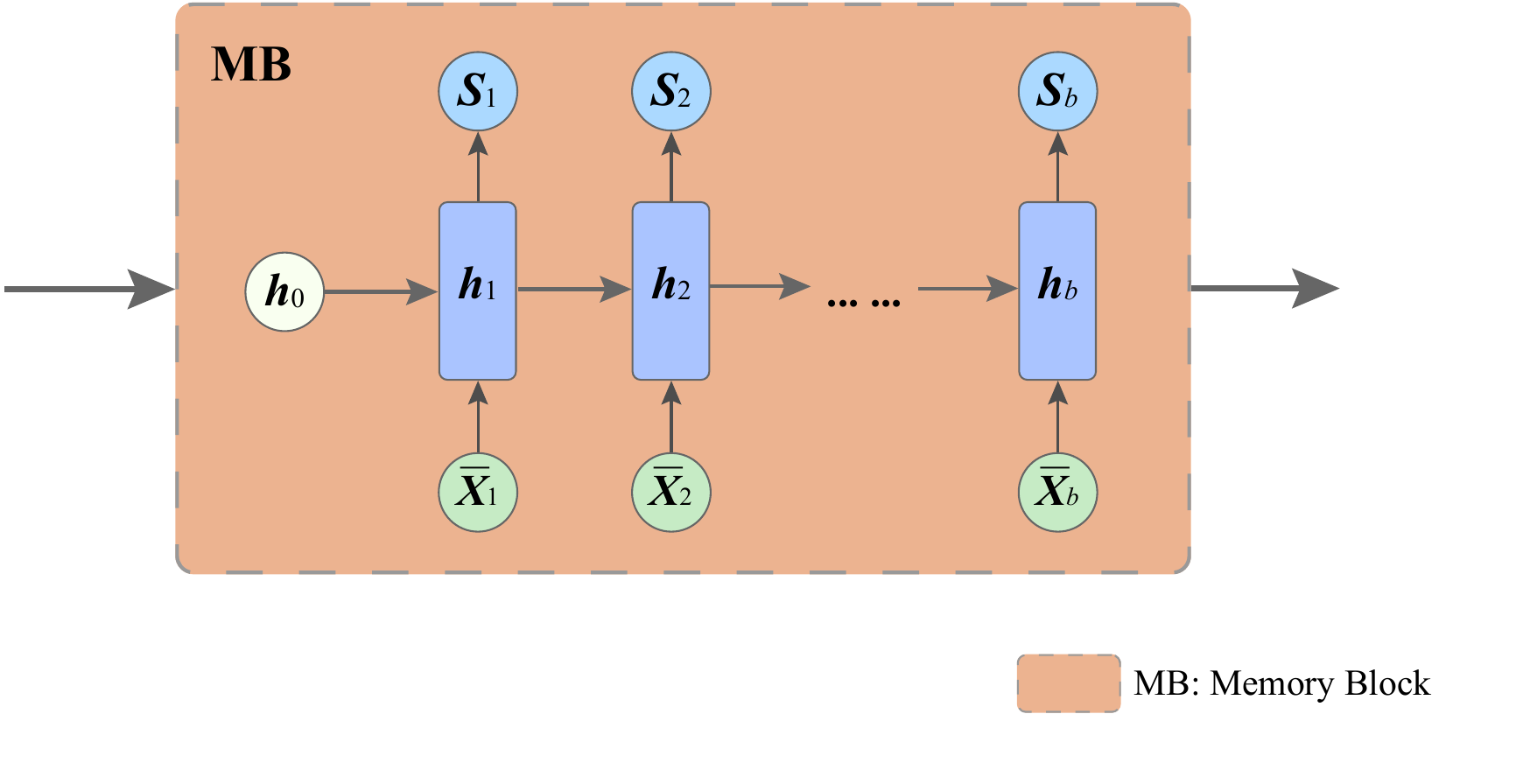}
\caption{Memory block. There are $b$ loops, indicating the memory block loops $b$ times, where $b$ is the number of batches. $\boldsymbol{h}_i$ represents the hidden layer's implicit information, i.e., the class prototypes information from the previous batch. ${{\overline{\boldsymbol{X}}}}_{i}$ is the original class prototypes information of the current batch, and $\boldsymbol{S}_i$ represents the final class prototypes until current batch.}
\label{fig3}
\end{figure}

 The memory block is shown in Figure \ref{fig3}. The learning of transferring information $\boldsymbol{h}_i$ is expressed as
\begin{equation}
\label{formula_ex5}
\boldsymbol{h}_{i} = \sigma( f_{h}\left( \boldsymbol{h}_{i - 1} \right) + f_{x}({{\overline{\boldsymbol{X}}}}_{i}) ),
\end{equation}
where $f(\cdot)$ is a linear layer, and $f_h(\cdot)$ and $f_x(\cdot)$ correspond to the transformations of $\boldsymbol{h}_{i - 1}$ and ${{\overline{\boldsymbol{X}}}}_{i}$, $\sigma(\cdot)$ uses the Sigmoid activation function. After obtaining $\boldsymbol{h}_{i}$, the final class prototypes information needs to be outputted, i.e.,
\begin{equation}
\label{formula_ex6}
\boldsymbol{S}_{i} = \sigma\left( f_{o}\left( \boldsymbol{h}_{i} \right) \right).
\end{equation}
 Here, $f_o(\cdot)$ is the transformation of the output class prototypes, and $\sigma(\cdot)$ uses the Tanh activation function. After obtaining the final class prototypes, the similarity between each sample and the class prototypes can be calculated. Cosine similarity is used to calculate the similarity between the refined high-level representation $\boldsymbol{m}_i$ of each sample and each final class prototype $\boldsymbol{S}^c$, where $\boldsymbol{S}^c$ is the $\boldsymbol{S}_{i}$ corresponding to category $c$. The calculation formula is 
\begin{equation}
\label{formula_ex7}
\widehat{p_{ic}} = {\mathit{\cos}\left( {\boldsymbol{m}_{i},~\boldsymbol{S}^{c}} \right)} = \frac{\boldsymbol{m}_{i} \cdot \boldsymbol{S}^{c}}{\left\| \boldsymbol{m}_{i} \right\|\left\| \boldsymbol{S}^{c} \right\|},
\end{equation}
where $i \in \left\{ {1,\ldots,N} \right\}$, $N$ is the total number of samples, and $\widehat{p_{ic}}$ is the similarity between the $i$-th sample and the $c$-th class prototype. Before calculating the loss function, the predicted similarity of the samples is normalized by
\begin{equation}
\label{formula_ex8}
p_{ic} = Softmax\left( \widehat{p_{ic}} \right) = \frac{e^{\widehat{p_{ic}}}}{\sum\limits_{c = 1}^{C}e^{\widehat{p_{ic}}}},
\end{equation}
so that the sum of the probabilities of the samples predicting different classes is equal to 1. After obtaining the predicted probability of each sample, as in the ordinary classification neural network training, the ordinary cross-entropy loss calculation can be performed. The cross-entropy loss here implies the inner-class distance, that is, the distance between samples and the corresponding class prototypes, which is calculated by 
\begin{equation}
\label{formula_ex9}
L_{inner} = - \frac{1}{N}{\sum\limits_{i = 1}^{N}{\sum\limits_{c = 1}^{C}{y_{ic}{\mathit{\log}\left( p_{ic} \right)}}}},
\end{equation}
where $y_{ic}$ is a sign function. If the true category of sample $i$ is equal to $c$, then $ y_{ic}=1$, otherwise $y_{ic}=0$. The inter-class distance among class prototypes is computed by  
\begin{equation}
\label{formula_ex10}
L_{inter} = \frac{2}{C \times (C - 1)}{\sum\limits_{n = 1}^{C}{\sum\limits_{m = n + 1}^{C}{\mathit{\cos}\left( \boldsymbol{S}^{n},~\boldsymbol{S}^{m} \right)}}}.
\end{equation}
\subsection{Training algorithm and test}
Algorithm \ref{alg1} is SAMN's training algorithm, in which $blocknum$ is the number of blocks in the sample attention module, and $K$ is the total number of layers in the extraction module.  

\begin{algorithm}[H]
\caption{$\{\boldsymbol{W}^{*},b^{*}\} = \mathbf{S}\mathbf{A}\mathbf{M}\mathbf{N}\left( T,epoch,\eta,blocknum, K \right)$}
\label{alg:alg1}
\begin{algorithmic}
\STATE 
\STATE \textbf{Input:} $T=\left\{\left({\boldsymbol{x}_{i},y_{i}}\right)\middle|\boldsymbol{x}_{i}\in\boldsymbol{R}^{n},i \in\left\{{1,\ldots,N}\right\},y_{i}\in\left\{{1,\ldots,C}\right\}\right\},epoch,\eta,blocknum,K.$
\STATE \textbf{Output:} $\theta = \left\{ {\boldsymbol{W}^{*},b^{*}} \right\} = \left\{ {\boldsymbol{W}^{j},~\boldsymbol{b}^{j},1 \leq j \leq K} \right\} \cup \{ \boldsymbol{W}^{h},\boldsymbol{b}^{h},\boldsymbol{W}^{x},\boldsymbol{b}^{x},\boldsymbol{W}^{s},\boldsymbol{b}^{s} \}.$
\STATE \hspace{0.5cm}(1) Randomly Initialization $\boldsymbol{W}^{j} \approx 0,~\boldsymbol{b}^{j} \approx 0\boldsymbol{~}\left( {1 \leq j \leq K} \right),\boldsymbol{h}^{c} \approx 0, k_{0}=K-1.$
\STATE \hspace{0.5cm}(2) \hspace{0.5cm}\textbf{For} $n=1:epoch$
\STATE \hspace{0.5cm}(3) \hspace{1cm}\textbf{//Feature Extraction}
\STATE \hspace{0.5cm}(4) \hspace{1cm}$\boldsymbol{h}_{0} = \boldsymbol{x}.$
\STATE \hspace{0.5cm}(5) \hspace{1cm}Compute $\boldsymbol{h}_{j} = \sigma\left( {\boldsymbol{W}^{j}\boldsymbol{h}_{j - 1} + \boldsymbol{b}^{j}} \right),1 \leq j \leq k_{0}.$
\STATE \hspace{0.5cm}(6) \hspace{1cm}\textbf{//Sample Attention Module}
\STATE \hspace{0.5cm}(7) \hspace{1cm}$\boldsymbol{X}^{c} = \boldsymbol{h}_{k_{0}}^{c}$
\STATE \hspace{0.5cm}(8) \hspace{1cm}\textbf{For} $block = 1:~blocknum$
\STATE \hspace{0.5cm}(9) \hspace{1.5cm}$\boldsymbol{Q}^{c} = \boldsymbol{K}^{c} = \boldsymbol{V}^{c} = \boldsymbol{X}^{c}.$
\STATE \hspace{0.5cm}(10) \hspace{1.35cm}Compute $\boldsymbol{A}^{c} = Softmax\left( {\boldsymbol{Q}^{c} \times \boldsymbol{K}^{c}} \right).$
\STATE \hspace{0.5cm}(11) \hspace{1.35cm}Compute $\boldsymbol{X}^{c} = \boldsymbol{A}^{c} \times \boldsymbol{V}^{c}.$
\STATE \hspace{0.5cm}(12) \hspace{0.85cm}\textbf{End For}
\STATE \hspace{0.5cm}(13) \hspace{0.85cm}\textbf{//Memory Block}
\STATE \hspace{0.5cm}(14) \hspace{0.85cm}Compute ${{\overline{\boldsymbol{X}}}}^{c} = mean\left( \boldsymbol{X}^{c} \right).$
\STATE \hspace{0.5cm}(15) \hspace{0.85cm}Compute $\boldsymbol{h}^{c} = \sigma(( \boldsymbol{W}^{h}\boldsymbol{h}^{c} + \boldsymbol{b}^{h} ) + (\boldsymbol{W}^{x}{{{\overline{\boldsymbol{X}}}}^{c}} + \boldsymbol{b}^{x} )).$
\STATE \hspace{0.5cm}(16) \hspace{0.85cm}Compute $\boldsymbol{S}^{c} = \sigma\left( {\boldsymbol{W}^{s}\boldsymbol{h}^{c} + \boldsymbol{b}^{s}} \right).$
\STATE \hspace{0.5cm}(17) \hspace{0.85cm}\textbf{//Further Extraction}
\STATE \hspace{0.5cm}(18) \hspace{0.85cm}Compute $\boldsymbol{h}_{j} = \sigma\left( {\boldsymbol{W}^{j}\boldsymbol{h}_{j - 1} + \boldsymbol{b}^{j}} \right),k_{0} \textless j \leq K.$
\STATE \hspace{0.5cm}(19) \hspace{0.85cm}\textbf{//Inner Class}
\STATE \hspace{0.5cm}(20) \hspace{0.85cm}$\boldsymbol{m} = \boldsymbol{h}_{K}$
\STATE \hspace{0.5cm}(21) \hspace{0.85cm}Compute $L_{inner} = - \frac{1}{N}{\sum\limits_{i = 1}^{N}{\sum\limits_{c = 1}^{C}{y_{ic}{\mathit{\log}{\left( {\mathit{Softmax}\left( {\mathit{\cos}\left( {\boldsymbol{m}_{i},~\boldsymbol{S}^{c}} \right)} \right)} \right).}}}}}$
\STATE \hspace{0.5cm}(22) \hspace{0.85cm}\textbf{//Inter Class}
\STATE \hspace{0.5cm}(23) \hspace{0.85cm}Compute $L_{inter} = \frac{2}{C \times (C - 1)}{\sum\limits_{n = 1}^{C}{\sum\limits_{m = n + 1}^{C}{\mathit{\cos}\left( \boldsymbol{S}^{n},~\boldsymbol{S}^{m} \right)}}}.$
\STATE \hspace{0.5cm}(24) \hspace{0.85cm}Compute $L_{total} = L_{inner} + L_{inter}.$
\STATE \hspace{0.5cm}(25) \hspace{0.85cm}Update weights and biases by back propagation with learning rate $\eta$.
\STATE \hspace{0.5cm}(26) \hspace{0.35cm}\textbf{End For}
\STATE \hspace{0.5cm}(27) \hspace{0.35cm}\textbf{Return}
\end{algorithmic}
\label{alg1}
\end{algorithm}

After SAMN is trained, the test process of a sample $\boldsymbol{x}$ is as follows: the similarities between $\boldsymbol{x}$ and all class prototypes are calculated, and the class prototype with the highest similarity is then considered as the category of $\boldsymbol{x}$.
\section{Experiments}
\label{experiments}
SAMN is expected to perform well on the classification tasks that need to take into account sample attention. Considering that tabular data may be the most appropriate scenario where sample attention is needed, we conduct experiments on 16 tabular datasets, including 10 binary classification datasets and 6 multi-classification datasets. To investigate the performance of SAMN compared to a single SVM or NN,  as well as existing methods combining SVM and NN, we select the famous SVC, the fully connected NN with the cross-entropy loss (CENet), and DNMSVM as baseline models. We also perform ablation studies on modules of SAMN to understand the effectiveness of sample attention and memory block.

\subsection{Datasets}
The experimental datasets come from LIBSVM Datasets \cite{csie} and UCI Datasets \cite{uci}. The data consists of natural and artificial datasets, with a wide range of variations in feature dimensions from 3 to 112 and sample sizes from 195 to 15,3000. These diverse datasets are commonly used in machine learning to evaluate model performance. All datasets were normalized before the experiments, with a mean of 0 and a standard deviation of 1. 

\subsection{Baseline models}
\paragraph{SVC:}Since we draw sample attention from the computing perspective of SVM, the most common SVC is selected for comparison. The model is implemented based on LIBSVM \cite{chang2011libsvm}.
\paragraph{CENet:} To compare with ordinary neural networks, we choose the most general structure of the fully connected network, using the cross-entropy loss function.
\paragraph{DNMSVM:}We also compare the most general research on the combination of SVM and neural network, DNMSVM. The network configuration for DNMSVM is based on the paper \cite{li2017deep}.

\subsection{Training scheme}
The data is split into training and testing sets using an 8:2 ratio. Then, the training set is further divided into a validation set, which consists of 20\% of the training data. To save time, a 5:5 ratio is adopted for the last two datasets with large sample sizes in multi-class classification, and 20\% of the data from the training set is also used as the validation set. The data is randomly divided five times, and the mean and standard deviation metrics are reported. For SVC, grid search with five-fold cross-validation is used to select hyperparameters, and the parameter range is $\gamma \in \{2^{-15}, 2^{-13}, \cdots, 2^{3}\}$ and $C\in\{2^{-5}, 2^{-3}, \cdots, 2^{15}\}$, according to the LIBSVM guide \cite{hsu2003practical}. Following the approach of Li et al. \cite{li2017deep}, a three-layer network structure is used for all networks, which is relatively stable. According to the recommendation of Bengio et al. \cite{bengio2012practical}, the number of hidden neurons in the network is set to the size of the input data dimension. The model parameters of CENet, DNMSVM, and SAMN are in the same order of magnitude. To avoid pairwise computation among class prototypes in multi-classification tasks, hidden layer $\boldsymbol{h}_{s} = \sigma\left( f_{s}\left( \boldsymbol{S} \right) \right)$ is used instead of cosine similarity. SAMN uses only one attention block, with $blocknum$ set to 1 and feature extraction layer $K$ set to 3. The Adam optimizer is used as the learning algorithm during training, with an initial learning rate $\eta$ of 0.01 and a maximum $epoch$ of 1000. The loss of all models converges before reaching the maximum number of iterations.

\subsection{Results}
\begin{table}[htbp]
  \centering
  \caption{Accuracy (in \%) and the standard deviation on  10 binary classification datasets. SAN is SAMN without memory block, and MBN is SAMN without sample attention.}
    \begin{tabular}{p{5.665em}p{4.165em}p{4.165em}p{4.165em}p{4.165em}p{4.165em}p{4.165em}}
    \toprule
    Dataset\textbackslash{}Model & \multicolumn{1}{c}{SVC} & \multicolumn{1}{c}{CENet} & \multicolumn{1}{c}{DNMSVM} & \multicolumn{1}{c}{SAN} & \multicolumn{1}{c}{MBN}  &  \multicolumn{1}{c}{SAMN} \\
    \midrule
    Parkinsons & 90.77±3.84 & 87.69±2.51 & 90.26±1.92 & 87.69±3.77 & 91.79±2.51 & \textbf{92.82±3.40} \\
    Sonar  & 84.28±2.86 & 79.52±3.87 & 82.38±2.43 & 84.29±7.47 & \textbf{85.24±3.16} & \textbf{85.24±3.16} \\
    Spectf   & 76.67±2.77 & 77.04±2.51 & 77.04±3.99 & 79.63±4.68 & 77.78±4.97 & \textbf{81.11±3.78} \\
    Heart   & 84.07±3.44 & 79.63±2.62 & 82.59±2.51 & 81.48±1.17 & 81.85±2.96 & \textbf{84.44±3.01} \\
    Ionosphere   & 92.11±3.94 & 91.27±2.87 & 86.48±3.84 & 92.96±3.45 & 93.52±4.04 & \textbf{94.08±3.82} \\
    Breast   & 97.37±1.57 & 95.09±1.19 & 96.67±1.51 & 96.67±1.70 & 97.02±1.97 & \textbf{97.89±1.19} \\
    Australian   & 86.23±1.21 & 83.48±2.12 & 85.65±3.79 & 86.81±2.12 & \textbf{87.97±1.75} & 86.96±2.05 \\
    German   & 73.50±2.51 & 69.9±2.06 & 74.1±2.67 & 73.7±2.54 & 73.0±2.70 & \textbf{75.1±4.02} \\
    Mushrooms   & 99.95±0.06 & 99.93±0.06 & 99.94±0.04 & \textbf{99.98±0.05} & \textbf{99.98±0.05} & \textbf{99.98±0.05} \\
    Phishing  & \textbf{97.48±0.20} & 96.23±0.52 & 96.09±0.54 & 95.74±0.22 & 96.83±0.24 & 96.93±0.53\\
    \bottomrule
    \end{tabular}%
  \label{tab:result1}%
\end{table}%
Table \ref{tab:result1} lists the results of various models on the binary classification datasets, in which SAM indicates SAMN without memory block while MBN is the SAMN without sample attention. The best results are marked in bold font. The results indicate SAMN achieves the best performance, and sample attention and memory block are indispensable.  SAMN obtains the highest test accuracy on eight datasets, MBN wins on three datasets, SAN and SVC win on one dataset, while CENet and DNMSVM do not win on any dataset. 
\begin{table}[htbp]
  \centering
  \caption{Accuracy, Precision, Recall, F1-score (in \%) and the standard deviation on 6 multi-classification datasets.}
    \begin{tabular}{lp{3.8em}p{5.2em}p{5.2em}p{5.2em}p{5.2em}}
    \toprule
    \multicolumn{1}{p{5.61em}}{Dataset} & Model & Accuracy & Precision & Recall & F1-score \\
    \midrule
    \multicolumn{1}{l}{\multirow{2}[2]{*}{Iris}} & CENet & 92.67±3.27 & 92.95±3.41 & 92.67±3.27 & 92.65±3.26 \\
          & SAMN  & \textbf{94.0±2.49} & \textbf{94.58±1.98} & \textbf{94.0±2.49} & \textbf{93.95±2.57} \\
  
    \multicolumn{1}{l}{\multirow{2}[2]{*}{Wine}} & CENet & 95.56±2.83 & 96.24±2.36 & 95.43±2.74 & 95.63±2.72 \\
          & SAMN  & \textbf{98.89±2.22} & \textbf{99.17±1.67} & \textbf{98.78±2.44} & \textbf{98.91±2.17} \\
   
    \multicolumn{1}{l}{\multirow{2}[2]{*}{Wine quality red}} & CENet & 59.12±2.15 & \textbf{34.06±5.62} & 29.03±1.30 & 29.51±1.23 \\
          & SAMN  & \textbf{63.0±1.53} & 32.13±3.10 & \textbf{31.15±2.99} & \textbf{31.04±2.97} \\
   
    \multicolumn{1}{l}{\multirow{2}[2]{*}{Dry bean}} & CENet & 92.34±0.20 & 93.57±0.28 & 93.36±0.13 & 93.45±0.19 \\
          & SAMN  & \textbf{92.55±0.20} & \textbf{93.84±0.17} & \textbf{93.53±0.24} & \textbf{93.67±0.19} \\
  
    \multicolumn{1}{l}{\multirow{2}[2]{*}{Batteryless sensor}} & CENet & 97.79±0.13 & 94.77±0.77 & \textbf{88.03±0.51} & 90.37±0.47 \\
          & SAMN  & \textbf{97.95±0.10} & \textbf{95.92±0.23} & 87.85±0.67 & \textbf{90.47±0.63} \\

    \multicolumn{1}{l}{\multirow{2}[2]{*}{Accelerometer}} & CENet & 58.44±0.92 & \textbf{60.0±2.05} & 58.44±0.92 & 56.16±2.34 \\
          & SAMN  & \textbf{58.67±0.99} & 58.44±1.98 & \textbf{58.67±0.99} & \textbf{57.37±0.80} \\
    \bottomrule
    \end{tabular}%
  \label{tab:result2}%
\end{table}%

Table \ref{tab:result2} presents the results of four metrics on the multi-classification datasets. In binary classification tasks, accuracy is commonly used as an evaluation metric. However, in multi-classification tasks, it is important to evaluate the model's performance across multiple categories. Therefore, precision, recall, and F1-score can be added to provide a more comprehensive evaluation of the model's performance. Since SVC and DNMSVM are designed for binary classification tasks, they need to be converted into binary classification problems to process multi-classification tasks. Therefore, we will not compare them there. From Table \ref{tab:result2}, we see that SAMN completely outperforms CENet in metrics of accuracy and F1-score. Although SAMN's precision and recall may be slightly lower than CENet's on some datasets, the precision and recall of SAMN are still superior to that of CENet on most datasets. It indicates that SAMN performs better on multi-classification tasks than CENet does. 
\section{Conclusion and future work}
\label{conclusion}
We propose a sample attention memory network (SAMN) that achieves a complementary effect between SVM and NN, which can be uniformly trained using the BP algorithm. SAMN has more capabilities than a single SVM or a single NN. It consists of sample attention module, class prototypes, memory block, as well as regular NN units. The function of SVM is realized by the sample attention module, class prototypes, and memory block, while the function of NN is realized by regular NN units. Extensive experiments show that SAMN achieves better classification performance than a single SVM or a single NN with a similar parameter size, as well as the previous best model in combining SVM and NN. The sample attention mechanism is a flexible block. It can be easily stacked, deepened, and incorporated into the neural networks that need it. 

Currently, we have verified the performance of SAMN on tabular datasets. It is a future work verifying SAMN on large and more diverse datasets. In addition, we will design deep SAMN in future work to enhance the model’s expression ability.

\bibliographystyle{unsrt}
\bibliography{references}


\end{document}